\documentclass[conference]{IEEEtran}
\IEEEoverridecommandlockouts
\usepackage{cite}
\usepackage{amsmath,amssymb,amsfonts,bm} 
\usepackage{algorithmic}
\usepackage{graphicx}
\usepackage{textcomp}
\usepackage{xcolor}
\usepackage{booktabs} 
\usepackage{url}
\usepackage[ruled,vlined]{algorithm2e} 
\usepackage{tikz}
\usetikzlibrary{shapes, arrows}
\usepackage{colortbl}
\usepackage{multirow}
\usepackage{subcaption}

\usetikzlibrary{shapes.geometric, arrows.meta, positioning, calc, shadows}

\graphicspath{{./}} 

\def\BibTeX{{\rm B\kern-.05em{\sc i\kern-.025em b}\kern-.08em
    T\kern-.1667em\lower.7ex\hbox{E}\kern-.125emX}}

\begin{document}

\title{Robust Path Tracking for Vehicles via Continuous-Time Residual Learning: An ICODE-MPPI Approach}

\author{Shugen Song$^{1}$, Wenjie Mei$^{2,\dagger}$, and Chengyan Zhao$^{3}$
\thanks{*This work is supported by the National Natural Science Foundation of China (NSFC)  under grant 62403125, the Natural Science Foundation of Jiangsu Province, and the Fundamental Research Funds for the Central Universities under grants 2242024k30037 and 2242024k30038. }
\thanks{$^{1}$Shugen Song is School of Automation,
        Southeast University, Nanjing 210096, China
        {\tt\small 13955012259@163.com}}%
\thanks{$^{2}$Wenjie Mei was with the School of Automation and the Key Laboratory of MCCSE of the Ministry of Education, Southeast University, Nanjing 210096, China, during the development of this work 
        {\tt\small mei.wenjie@nju.edu.cn}}%
\thanks{$^{3}$Chengyan Zhao is with Graduate School of Computer Science and Systems Engineering, Kyushu Institute of Technology, 680-4 Kawazu, Iizuka-shi, Fukuoka, Japan 
        {\tt\small zcy\_neu@hotmail.com}}%
\thanks{$^{\dagger}$Corresponding author.}
}

\maketitle

\begin{abstract}
Model Predictive Path Integral (MPPI) control is a powerful sampling-based strategy for nonlinear autonomous systems. However, its performance is often bottlenecked by the fidelity of nominal dynamics. We propose ICODE-MPPI, a robust framework that leverages Input Concomitant Neural Ordinary Differential Equations (ICODEs) to learn and compensate for unmodeled residual dynamics. Unlike discrete-time learners, ICODEs maintain physical consistency and temporal continuity during the MPPI prediction horizon. High-fidelity simulations on complex trajectories demonstrate that ICODE-MPPI achieves up to a 69\% reduction in cross-tracking error under persistent disturbances compared to standard MPPI control. Furthermore, our analysis confirms that ICODE-MPPI significantly suppresses control chattering, yielding smoother steering commands and superior robust performance.
\end{abstract}


\section{Introduction}
Precise path tracking and proactive obstacle avoidance for autonomous ground vehicles (AGVs) are fundamental to intelligent transportation systems, particularly in unstructured or disturbed environments \cite{testouri2023towards}. Model Predictive Control (MPC) has long been the paradigm for such tasks \cite{zhang2025model}; however, its gradient-based nature often struggles with non-convex cost landscapes. Recently, Model Predictive Path Integral (MPPI) control, a derivative-free and sampling-based variant, has gained prominence for its ability to handle complex constraints and nonlinear dynamics \cite{williams2016aggressive}. Its universality has been demonstrated in agile UAV maneuvers \cite{minavrik2024model}, USV path-following \cite{liu2022data}, and quadrotor pursuit \cite{jeong2022mppi}.

Despite its robustness, the efficacy of MPPI is fundamentally sensitive to model-plant mismatch. Since the control law is derived from an expectation over sampled trajectories, inaccuracies in the forward dynamics model lead to biased rollouts, resulting in suboptimal control sequences, persistent tracking offsets, or high-frequency chattering \cite{ross2012agnostic, kim2022smooth}. These challenges have necessitated a change toward data-driven control (DDC) to capture complex uncertainties more effectively \cite{hou2013model}.

Early data-driven extensions utilized multi-layer perceptrons (MLPs) \cite{nagabandi2018neural} or Gaussian processes (GPs) \cite{kabzan2019learning}. However, MLPs often lack physical consistency in discrete-time transitions, while GPs suffer from prohibitive computational overhead in real-time optimization. Neural ordinary differential equations (Neural ODEs) \cite{chen2018neural} address these by parameterizing the state derivative, providing a continuous-time representation. For safety-critical control, architectures like ControlSynth ODEs (CSODEs) \cite{mei2024controlsynth} and input concomitant neural ODEs (ICODEs) \cite{li2025icode} further introduce convergence/contraction-guaranteed stability, demonstrating superior fidelity in modeling, for example, mechanical and electromechanical systems \cite{mei2025learning}.

While many efforts have explored Transformers \cite{zinage2024transformer} or Reinforcement Learning \cite{qu2023rl} to enhance MPPI, integrating continuous-time, stability-guaranteed methods with its stochastic sampling architecture remains an open challenge. Inspired by residual learning \cite{zhang2025model} and robust sampling-based control \cite{gandhi2021robust}, we propose the ICODE-MPPI framework. By employing ICODE as a continuous-time residual structure, our approach preserves the control-affine structure of the system while ensuring that high-frequency samples remain physically consistent and robust to disturbances.

The main contributions of this work are:
\begin{itemize}
\item We develop a residual dynamics framework based on ICODEs and iterative data aggregation. This hybrid strategy captures continuous-time uncertainties while maintaining high physical consistency.
\item We integrate the trained ICODE model into the MPPI prediction horizon, enabling proactive compensation for environmental disturbances and significantly mitigating the steady-state tracking drift inherent in nominal models.
\item We provide a comprehensive benchmarking of ICODE-MPPI against baselines, evaluating state-variable RMSE and control rate probability densities to demonstrate superior tracking precision and control smoothness.
\end{itemize}

The remainder of this paper is organized as follows. Section II presents the problem formulation and preliminaries. Section III details the proposed approach. Section IV discusses simulation and results, and Section V concludes the paper.

\section{ Preliminaries and Problem Formulation}
\label{sec:problem_formulation}

In this section, we formally present the stochastic optimal control problem and detail the vehicle kinematics and MPPI algorithm used for trajectory planning.

\subsection{Problem Formulation}
Consider a general discrete-time nonlinear dynamical system derived from a continuous-time process. The system evolution is governed by the following equation:
\begin{equation}
    \mathbf{x}_{t+1} = \mathbf{F}(\mathbf{x}_t, \mathbf{v}_t)
    \label{eq:system_dynamics}
\end{equation}
where $\mathbf{x}_t \in \mathbb{R}^{n_x}$ denotes the system state vector, and $\mathbf{v}_t \in \mathbb{R}^{n_u}$ represents the noisy control input. The control input is modeled as the nominal control $\mathbf{u}_t$ perturbed by Gaussian noise:
\begin{equation}
    \mathbf{v}_t = \mathbf{u}_t + \bm{\epsilon}_t, \quad \bm{\epsilon}_t \sim \mathcal{N}(0, \Sigma)
    \label{eq:noise_model}
\end{equation}
where $\Sigma$ is the covariance matrix representing the exploration variance.

The objective is to find an optimal control sequence $U^* = \{\mathbf{u}_0, \mathbf{u}_1, \dots, \mathbf{u}_{H-1}\}$ that minimizes the expected cost over a finite horizon $H$. Formally, this optimization problem is formulated as follows:
\begin{equation}
\begin{aligned}
    \min_{U} \quad & J(U) = \mathbb{E}_{\mathbb{Q}} \left[ \phi(\mathbf{x}_H) + \sum_{t=0}^{H-1} \left( s(\mathbf{x}_t) + \frac{1}{2} \mathbf{u}_t^\top R \mathbf{u}_t \right) \right] \\
    \textrm{s.t.} \quad & \mathbf{x}_{t+1} = \mathbf{F}(\mathbf{x}_t, \mathbf{v}_t) \\
    & \mathbf{v}_t = \mathbf{u}_t + \bm{\epsilon}_t, \quad \bm{\epsilon}_t \sim \mathcal{N}(0, \Sigma) \\
    & h(\mathbf{x}_t, \mathbf{u}_t) \leq 0 \\
    & \mathbf{x}_0 = \mathbf{x}_{ini}
\end{aligned}
\label{eq:optimization_problem}
\end{equation}
where $\phi(\mathbf{x}_H)$ is the terminal cost, and the running cost is decomposed into a state-dependent cost $s(\mathbf{x}_t)$ and a quadratic control cost weighted by matrix $R$. The inequality $h(\mathbf{x}_t, \mathbf{u}_t) \leq 0$ represents physical constraints.

\subsection{Vehicle Dynamics Modeling}
For simplicity and to reduce computational complexity, this
work adopts a bicycle model to represent the motion model of
the vehicle \cite{testouri2023towards}.The state vector is defined as $\mathbf{x} = [x, y, \theta, v, \delta]^\top$, where $(x, y)$ is the global position, $\theta$ is the heading angle, $v$ is the longitudinal velocity, and $\delta$ is the steering angle. The control input vector is $\mathbf{u} = [a, \omega]^\top$, consisting of the longitudinal acceleration $a$ and the steering rate $\omega$.

The continuous-time dynamics are described by
\begin{equation}
    \frac{d}{dt} \begin{pmatrix} x \\ y \\ \theta \\ v \\ \delta \end{pmatrix} = \begin{pmatrix} v \cos(\theta) \\ v \sin(\theta) \\ \frac{v}{l} \tan(\delta) \\ a \\ \omega \end{pmatrix}
    \label{eq:nominal model}
\end{equation}
where $l$ is the wheelbase of the vehicle. This model captures the non-holonomic constraints of the vehicle without explicitly modeling the sideslip angle, suitable for trajectory planning at moderate speeds.

The vehicle motion model described by \eqref{eq:nominal model} can be rewritten in the general form of $\mathbf{x}_{t+1} = \mathbf{F}(\mathbf{x}_t, \mathbf{v}_t)$ by discretization, serving as the vehicle nominal model.

\subsection{Model Predictive Path Integral Algorithm}
MPPI solves the stochastic optimal control problem \eqref{eq:optimization_problem} by sampling trajectories, which has shown superior performance in handling complex nonlinear dynamics \cite{williams2016aggressive}. The algorithm consists of the following steps:

\subsubsection{Trajectory Sampling}
We sample $K$ rollout trajectories by perturbing the nominal control sequence $\mathbf{u}_t$. The perturbed control input for the $k$-th sample at time step $t$ is calculated by
\begin{equation}
    \mathbf{v}_t^k = \mathbf{u}_t + \bm{\epsilon}_t^k, \quad \bm{\epsilon}_t^k \sim \mathcal{N}(0, \Sigma)
    \label{eq:mppi_sampling}
\end{equation}
The system is forward simulated using the dynamics model to obtain state trajectories $\tau_k = \{\mathbf{x}_0^k, \dots, \mathbf{x}_H^k\}$.

\subsubsection{Cost Evaluation}
For each trajectory $\tau_k$, the total cost $S(\tau_k)$ is evaluated based on the running cost $q(\mathbf{x}_t, \mathbf{u}_t)$ and the terminal cost $\phi(\mathbf{x}_H)$:
\begin{equation}
    S(\tau_k) = \phi(\mathbf{x}_H^k) + \sum_{t=0}^{H-1} q(\mathbf{x}_t^k, \mathbf{u}_t)
    \label{eq:mppi_cost}
\end{equation}
Note that the control cost is implicitly handled in the weighting step through the change of measurement method in MPPI control.

\subsubsection{Weight Calculation}
The importance weight $w_k$ for the $k$-th trajectory is computed via the softmax function:
\begin{equation}
    w_k = \frac{1}{\eta} \exp \left( -\frac{1}{\lambda} \left( S(\tau_k) - \rho \right) \right)
    \label{eq:mppi_weights}
\end{equation}
where $\lambda$ is the temperature parameter, and $\rho = \min_{k} S(\tau_k)$ is a constant for numerical stability. The normalization factor $\eta$ is given by:
\begin{equation}
    \eta = \sum_{j=1}^{K} \exp \left( -\frac{1}{\lambda} \left( S(\tau_j) - \rho \right) \right)
    \label{eq:mppi_eta}
\end{equation}

\subsubsection{Control Update}
Finally, the resulting control sequence is updated by the probability-weighted average of the sampled noise:
\begin{equation}
    \mathbf{u}_t^* = \mathbf{u}_t + \sum_{k=1}^{K} w_k \bm{\epsilon}_t^k
    \label{eq:mppi_update}
\end{equation}
Furthermore, the obtained control sequence is then smoothed using a Savitzky-Golay filter. This formulation enables MPPI to approximate the optimal control by picking up low-cost trajectories while still maintaining exploration through the stochastic sampling process.

\section{Proposed approach}
\label{sec:methodology}

In this section, we present the proposed framework for robust trajectory tracking. We first detail the architecture of the Input Concomitant Neural Ordinary Differential Equations (ICODE) used to model the residual dynamics. Then, we describe the integration of ICODE with the MPPI controller, illustrated by a system flowchart.Finally, we discuss the data collection strategy and the training procedure.

\subsection{ICODE Architecture for Residual Modeling}
Standard Neural ODEs \cite{chen2018neural} typically model the derivative of the state as an autonomous function $\dot{\mathbf{x}}(t) = f_\theta(\mathbf{x}(t))$, treating external inputs as unknown parameters. However, for robotic systems, the state evolution is explicitly driven by control inputs. To accurately capture this interaction, we employ the ICODE architecture proposed in \cite{li2025icode}.

Unlike purely black-box models that concatenate state and control action into a single input vector, ICODE imposes a control-affine prior on the learned dynamics. This inductive bias ensures that the residual term aligns with the fundamental structure of robotic systems, where control inputs enter the state derivatives linearly, thereby improving sample efficiency. 

We present the learned ICODE dynamics $\mathbf{f}_{res}(\mathbf{x}, \mathbf{u}; \theta)$ as:
\begin{equation}
    \dot{\mathbf{x}}_{res}(t) = \mathbf{f}_{res}(\mathbf{x}(t), \mathbf{u}(t); \theta) = \mathbf{f}_\theta(\mathbf{x}(t)) + \sum_{j=1}^{m} \mathbf{g}_{\theta, j}(\mathbf{x}(t)) u_j(t)
    \label{eq:icode_structure}
\end{equation}
where $\mathbf{u}(t) = [u_1, \dots, u_m]^\top$ is the control input vector.
\begin{itemize}
    \item $\mathbf{f}_\theta(\mathbf{x}) \colon \mathbb{R}^{n_x} \to \mathbb{R}^{n_x}$ represents the state-dependent autonomous drift (e.g., unmodeled friction, aerodynamic drag).
    \item $\mathbf{g}_{\theta, j}(\mathbf{x})\colon \mathbb{R}^{n_x} \to \mathbb{R}^{n_x}$ represents the control-dependent gain fields, capturing how the $j$-th control input influences the state derivative (e.g., actuator effectiveness).
\end{itemize}

This affine decomposition, consistent with Eq. (2) in \cite{li2025icode}, provides a strong inductive bias. It ensures that the learned model respects the superposition principle of control inputs locally, significantly improving sample efficiency and generalization compared to fully connected networks. The total predicted dynamics used for MPPI rollouts combine the vehicle nominal model and the learned ICODE residual:
\begin{equation}
    \dot{\mathbf{x}}_{pred}(t) = \mathbf{f}_{nom}(\mathbf{x}(t), \mathbf{u}(t)) + \mathbf{f}_{res}(\mathbf{x}(t), \mathbf{u}(t); \theta)
    \label{eq:combined_dynamics}
\end{equation}

\subsection{ICODE-MPPI Integration Algorithm}
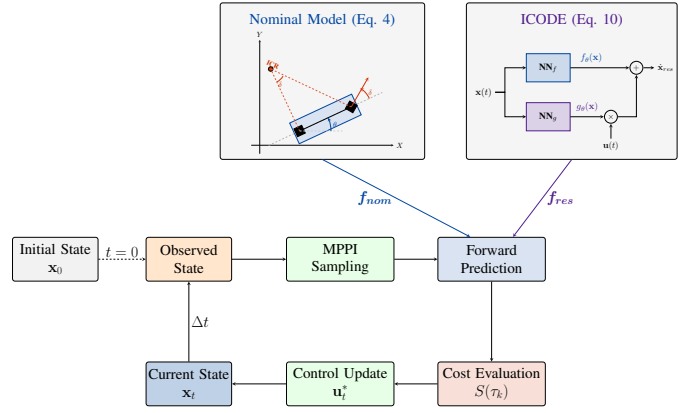
\begin{figure}[htbp]
    \centering
    \resizebox{1.0\columnwidth}{!}{
    \begin{tikzpicture}[>=stealth, auto, node distance=3.0cm]
      
        \definecolor{techblue}{RGB}{0, 70, 160}
        \definecolor{accentorange}{RGB}{210, 50, 0}
        \definecolor{neuralpurple}{RGB}{80, 30, 140}
        
        \tikzstyle{state} = [rectangle, draw, fill=techblue!25, very thick, text width=13em, text centered, rounded corners, minimum height=7em, font=\Huge]
        \tikzstyle{proc} = [rectangle, draw, fill=gray!25, very thick, text width=16.5em, text centered, rounded corners, minimum height=7em, font=\Huge]
        \tikzstyle{obs} = [rectangle, draw, fill=orange!20, very thick, text width=13em, text centered, rounded corners, minimum height=7em, font=\Huge]

        \tikzstyle{model_nom} = [rectangle, draw, fill=gray!8, text width=32em, text centered, rounded corners, minimum height=25em, font=\small]
        \tikzstyle{model_icode} = [rectangle, draw, fill=gray!8, text width=33em, text centered, rounded corners, minimum height=25em, font=\small]

        \tikzstyle{netF} = [rectangle, draw=techblue, fill=techblue!20, text width=4em, text centered, minimum height=3em, font=\bfseries]
        \tikzstyle{netG} = [rectangle, draw=neuralpurple, fill=neuralpurple!20, text width=4em, text centered, minimum height=3em, font=\bfseries]
        \tikzstyle{line} = [draw, ->, ultra thick] 
        \tikzstyle{dashline} = [draw, ->, ultra thick, dashed] 

        \node[state, fill=gray!10] (x_init) at (-6, 0) {Initial State \\ $\mathbf{x}_0$}; 
        \node[obs] (obs) at (1.5, 0) {Observed State};

        \node[proc, fill=green!10] (noise) at (10, 0) {MPPI \\ Sampling};
        \node[proc, fill=techblue!15] (pred) at (18.5, 0) {Forward \\ Prediction};

        \node[state] (x0) at (1.5, -7) {Current State \\ $\mathbf{x}_t$};
        \node[proc, fill=green!10] (update) at (10, -7) {Control Update \\ $\mathbf{u}^*_t$};
        \node[proc, fill=accentorange!15] (cost) at (18.5, -7) {Cost Evaluation \\ $S(\tau_k)$};

        \node[model_nom] (nom) at (9, 10.0) {};
        \node[anchor=north, font=\Huge, text=techblue, yshift=-0.5cm] at (nom.north) {Nominal Model (Eq. 4)};

        \begin{scope}[shift={(nom.center)}, yshift=-2.8cm, scale=1.6]
    
        \draw[->, black, thick] (-2.5,-0.5) -- (2.5,-0.5) node[right, scale=1.2] {$X$};
        \draw[->, black, thick] (-2.2,-0.8) -- (-2.2,3.2) node[above, scale=1.2] {$Y$};
    
        \begin{scope}[shift={(-0.8,0)}, rotate=25]
        \draw[fill=techblue!15, draw=techblue, very thick] (-0.2,-0.4) rectangle (2.2,0.4);
        \draw[gray!60, dashed, thin] (-1.2,0) -- (3.2,0); 
        \draw[ultra thick, black] (0,0) -- (2,0); 
        \draw[fill=black] (-0.15,-0.15) rectangle (0.15,0.15); 
        
        \begin{scope}[rotate=-25]
            \draw[gray!40, dashed, thick] (0,0) -- (1.5,0);
            \draw[techblue, ultra thick, ->] (1.1,0) arc (0:25:1.1) node[midway, right, scale=1.2, font=\bfseries] {$\theta$};
        \end{scope}

        \begin{scope}[shift={(2,0)}, rotate=35]
            \draw[fill=black] (-0.15,-0.15) rectangle (0.15,0.15);
            \draw[->, accentorange, ultra thick] (0,0) -- (1.2,0); 
        \end{scope}
        
        \draw[accentorange, ultra thick, ->] (2.7,0) arc (0:35:0.8) node[midway, right, scale=1.2] {$\delta$};
        
        \coordinate (ICR) at (0, 2.4);
        \draw[fill=accentorange] (ICR) circle (0.08) node[above, accentorange, scale=1.1, rotate=-25, font=\bfseries] {ICR};
        \draw[dashed, accentorange!80, thick] (0,0) -- (ICR);
        \draw[dashed, accentorange!80, thick] (2,0) -- (ICR);
        \draw[accentorange, semithick] (0, 1.9) arc (-90:-55:0.5) node[midway, below, scale=1.2] {$\delta$};
    \end{scope}
\end{scope}

        \node[model_icode] (icode) at (23, 10.0) {};
        \node[anchor=north, font=\Huge, text=neuralpurple, yshift=-0.5cm] at (icode.north) {ICODE (Eq. 10)};

        \begin{scope}[shift={(icode.center)}, yshift=-2cm, scale=1.1] 
    
        \node (ix) at (-4.5, 1.25) [scale=1.5] {$\mathbf{x}(t)$};
        \node (iu) at (2.0, -1.5) [scale=1.4] {$\mathbf{u}(t)$}; 
    
        \node[netF] (nf) at (-1.2, 2.5)[scale=1.5] {$\text{NN}_f$}; 
        \node[netG] (ng) at (-1.2, 0)[scale=1.5]{$\text{NN}_g$};
    
        \node[circle, draw, inner sep=2pt, ultra thick, scale=1.3] (mult) at (2.0, 0) {$\times$};
        \node[circle, draw, inner sep=2pt, ultra thick, scale=1.3] (plus) at (3.3, 2.5) {$+$};
    
        \node (f_val) at (1.0, 3.0) [scale=1.5, techblue] {$f_{\theta}(\mathbf{x})$};
        \node (g_val) at (0.8, 0.5) [scale=1.5, neuralpurple] {$g_{\theta}(\mathbf{x})$};
    
        \draw[line] (ix.east) -- ++(0.5,0) |- (nf.west);
        \draw[line] (ix.east) -- ++(0.5,0) |- (ng.west);
        \draw[line] (iu.north) -- (mult.south); 
        \draw[line] (ng.east) -- (mult.west);
        \draw[line] (nf.east) -- (plus.west); 
        \draw[line] (mult.east) -| (plus.south);
    
        \draw[line, ultra thick] (plus.east) -- ++(0.55,0) node[right, scale=1.4] {$\dot{\mathbf{x}}_{res}$};
    \end{scope}

       \draw[dashline] (x_init) -- node[above, font=\Huge] {$t=0$} (obs);
       \draw[line] (obs) -- (noise);
       \draw[line] (noise) -- (pred);
       \draw[line] (pred) -- (cost);
       \draw[line] (cost) -- (update);
       \draw[line] (update) -- (x0);
       \draw[line] (x0) -- node[midway, right, font=\Huge] {$\Delta t$} (obs);

        \draw[line, techblue, ultra thick] (nom.south) -- ([xshift=-1.2cm]pred.north) 
        node[midway, left, xshift=-0.1cm, font=\Huge] {$\bm{f_{nom}}$};
    
        \draw[line, neuralpurple, ultra thick] (icode.south) -- ([xshift=1.2cm]pred.north) 
        node[midway, right, xshift=0.1cm, font=\Huge] {$\bm{f_{res}}$};

    \end{tikzpicture}
    }
    \caption{Flowchart of the proposed ICODE-MPPI framework.}
    \label{fig:ICODE-MPPI framework}
\end{figure}

The proposed framework, as illustrated in Fig.~\ref{fig:ICODE-MPPI framework},integrates the learned ICODE model into the MPPI prediction horizon to bridge the gap between idealized modeling and real-world disturbances. Specifically, the system dynamics are reformulated as a composite structure in \eqref{eq:combined_dynamics}: the nominal model provides the fundamental physical prior as defined in \eqref{eq:nominal model},while the ICODE module is explicitly trained to capture the complex, state-dependent residual dynamics that are otherwise neglected or difficult to model accurately.

At each control step, the MPPI controller leverages this augmented model within its forward prediction module. By simulating thousands of potential trajectories using the integrated Nominal-ICODE dynamics, the controller can anticipate environmental disturbances and unmodeled forces in real-time. Compared to a baseline controller relying solely on the nominal model, this integration allows the ICODE-MPPI scheme to compensate for residual errors proactively, ensuring robust path tracking in highly disturbed environments.

\subsection{Data Collection and Iterative Training}
To ensure the learned model accurately represents the vehicle's dynamics during high-performance maneuvers, we implement an iterative data aggregation framework\cite{liu2022data}. This approach expands the dataset $\mathcal{D}$ by combining initial exploration with task-specific trajectories, addressing the distribution shift between random sampling and controlled path following.

\subsubsection{Data Sources and Aggregation}
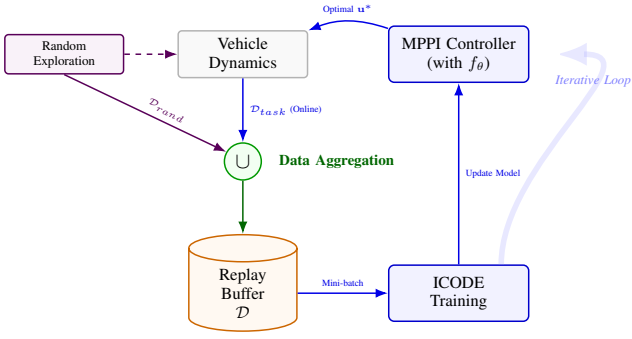
\begin{figure}[htbp]
\centering
\resizebox{\columnwidth}{!}{
\begin{tikzpicture}[
    node distance=1.2cm and 1.5cm,
    >=Latex,
    mainbox/.style={rectangle, draw=blue!80!black, fill=blue!5, thick, text width=2.4cm, text centered, minimum height=3em, rounded corners=3pt, font=\small},
    databox/.style={cylinder, draw=orange!80!black, fill=orange!5, thick, shape border rotate=90, aspect=0.25, text width=1.8cm, text centered, minimum height=3.5em, font=\small},
    procbox/.style={rectangle, draw=gray!60, fill=gray!5, thick, text width=2.2cm, text centered, minimum height=2.5em, rounded corners=2pt, font=\small},
    rand_style/.style={rectangle, draw=violet!70!black, fill=violet!5, thick, text width=2.0cm, text centered, font=\scriptsize, rounded corners=2pt},
    arrow/.style={->, thick, color=blue!90!black}
]

    \node [procbox] (env) {Vehicle Dynamics};
    \node [mainbox, right=1.5cm of env] (mppi) {MPPI Controller \\ (with $f_\theta$)};
    
    \node [rand_style, left=1cm of env] (rand) {Random \\ Exploration};
    
    \node [draw, circle, thick, fill=green!5, draw=green!60!black, below=1.2cm of env] (agg) {\large $\cup$};
    \node [right=0.2cm of agg, font=\footnotesize\bfseries, color=green!40!black] {Data Aggregation};

    \node [databox, below=1.0cm of agg] (buffer) {Replay Buffer \\ $\mathcal{D}$};
    \node [mainbox] (train) at (mppi |- buffer) {ICODE \\ Training};
 
    \draw [arrow, dashed, violet!70!black] (rand) -- (env);
    \draw [arrow, violet!70!black] (rand.south) -- node[above, sloped, font=\tiny, pos=0.6] {$\mathcal{D}_{rand}$} (agg.150);

    \draw [arrow] (mppi) edge[bend right=20] node[above, font=\tiny] {Optimal $\mathbf{u}^*$} (env);
    \draw [arrow] (env.south) -- node[right, font=\tiny] {$\mathcal{D}_{task}$ (Online)} (agg);

    \draw [arrow, thick, color=green!40!black] (agg) -- (buffer);

    \draw [arrow] (buffer) -- node[above, font=\tiny] {Mini-batch} (train);
    \draw [arrow] (train.north) -- node[right, font=\tiny] {Update Model} (mppi.south);

    \draw [->, blue!30, line width=3pt, opacity=0.3] ($(train.north)+(0.8,0)$) to [out=90, in=0, looseness=1.2] ($(mppi.east)+(0.5,0)$);
    \node [blue!50, font=\scriptsize\itshape] at ($(mppi.east)+(1.2, -0.5)$) {Iterative Loop};

\end{tikzpicture}
}
\caption{The iterative learning framework with data aggregation.}
\label{fig:data aggregation framework}
\end{figure}

The training dataset is constructed from two distinct data streams:
\begin{itemize}
    \item \textbf{Random Exploration ($\mathcal{D}_{rand}$):} The vehicle is initially excited with random control inputs to capture the broad physical response and stability boundaries across the operational envelope.
    \item \textbf{Task-Specific Data ($\mathcal{D}_{task}$):} As training progresses, the MPPI controller utilizes the current model $f_\theta$ to execute the path-following task. The resulting state-action transitions are collected as on-policy data.
\end{itemize}
As illustrated in Fig. \ref{fig:data aggregation framework}, these streams are aggregated such that $\mathcal{D} = \mathcal{D}_{rand} \cup \mathcal{D}_{task}$. This hybrid dataset ensures the model achieves high fidelity both in general dynamics and in the high-probability state regions near the reference trajectory.

\subsubsection{Training Procedure}
The neural network is optimized iteratively. In each cycle, a mini-batch $\mathcal{B}$ is sampled from the aggregated buffer $\mathcal{D}$ to minimize the multi-step prediction error. By integrating the nominal dynamics $\mathbf{f}_{nom}$ and the residual network $\mathbf{f}_{res}$ via an RK4 solver, the loss function is defined as:
\begin{equation}
    \mathcal{L}(\theta) = \frac{1}{|\mathcal{B}|} \sum_{i \in \mathcal{B}} \| \mathbf{x}_{t+1}^{(i)} - \Phi_{RK4}(\mathbf{x}_t^{(i)}, \mathbf{u}_t^{(i)}; \mathbf{f}_{nom}, \mathbf{f}_{res}) \|^2
\end{equation}
The updated model $f_\theta$ is then redeployed into the MPPI planner for the next round of data collection, forming a self-correcting feedback loop that continuously reduces modeling uncertainty.

\section{Simulations and Results}
\label{sec:experiments}

To evaluate the efficacy and robustness of the proposed ICODE-MPPI framework, we conduct high-fidelity simulations on a trajectory tracking task under unmodeled environmental disturbances. We conduct simulations on three diverse reference paths: \textbf{Ellipse}, \textbf{Sine-wave}, and \textbf{Figure-8}. These trajectories present increasing levels of difficulty in terms of curvature change and control agility. This section details the simulation environment, the disturbance modeling, and the performance metrics used for comparison.

\subsection{Experimental Setup}

\subsubsection{Dynamic Disturbance Environment}
The environment is characterized by persistent external disturbances $\bm{\delta}(t)$. Unlike the ideal nominal model, the actual state evolution follows $\dot{\mathbf{x}} = \mathbf{f}_{nom} + \bm{\delta}(t)$. As implemented in our simulation, the disturbance is a composite sinusoidal signal:
\begin{equation}
    \bm{\delta}(t) = [A_x \sin(\omega_x t), A_y \cos(\omega_y t), A_\theta \sin(\omega_\theta t), 0, 0]^\top
\end{equation}
This setup manifests as time-varying drifting in the $X, Y$ coordinates and the heading angle $\theta$, creating a significant model mismatch for any controller relying solely on the nominal model.

\subsubsection{Hyper-parameters}
The key parameters extracted from our implementation are summarized in Table \ref{tab:all_params}. These ensure a fair comparison between the baseline MPPI and our ICODE-MPPI.

\begin{table}[htbp]
    \caption{System Configurations and Hyper-parameters }
    \centering
    \renewcommand{\arraystretch}{1.2} 
    \begin{tabular}{llc} 
        \toprule
        \textbf{Category} & \textbf{Parameter} & \textbf{Value} \\
        \midrule
        
        \multirow{4}{*}{\shortstack{\textbf{Vehicle} \\ \textbf{Kinematics}}} 
        & Wheelbase ($L$) & 2.5 m \\
        & Max Acceleration ($a_{max}$) & 2.0 m/s$^2$ \\
        & Max Steering Angle ($\delta_{max}$) & 1.571 rad \\
        & Sampling Time ($\Delta t$) & 0.05 s \\
        \midrule 
        
        \multirow{4}{*}{\shortstack{\textbf{MPPI} \\ \textbf{Controller}}} 
        & Prediction Horizon ($H$) & 20 steps \\
        & Number of Samples ($K$) & 3000 \\
        & Temperature Parameter ($\lambda$) & 0.05 \\
        & Control Noise Covariance ($\Sigma$) & diag(0.1, 0.5) \\
        \midrule
        
        \multirow{5}{*}{\shortstack{\textbf{ICODE} \\ \textbf{Network}}} 
        & Hidden dimension & [256, 256]  \\
        & Number of Layers  & 3 \\
        & Activation Function & Softplus \\
        & Learning Rate & $5 \times 10^{-4}$ \\
        & Optimizer & Adam \\
        \bottomrule
    \end{tabular}
    \label{tab:all_params}
\end{table}

\subsection{Trajectory Tracking and Qualitative Analysis}
\begin{figure*}[t]
    \centering
    \begin{subfigure}[b]{0.32\textwidth}
        \centering
        \includegraphics[width=\textwidth]{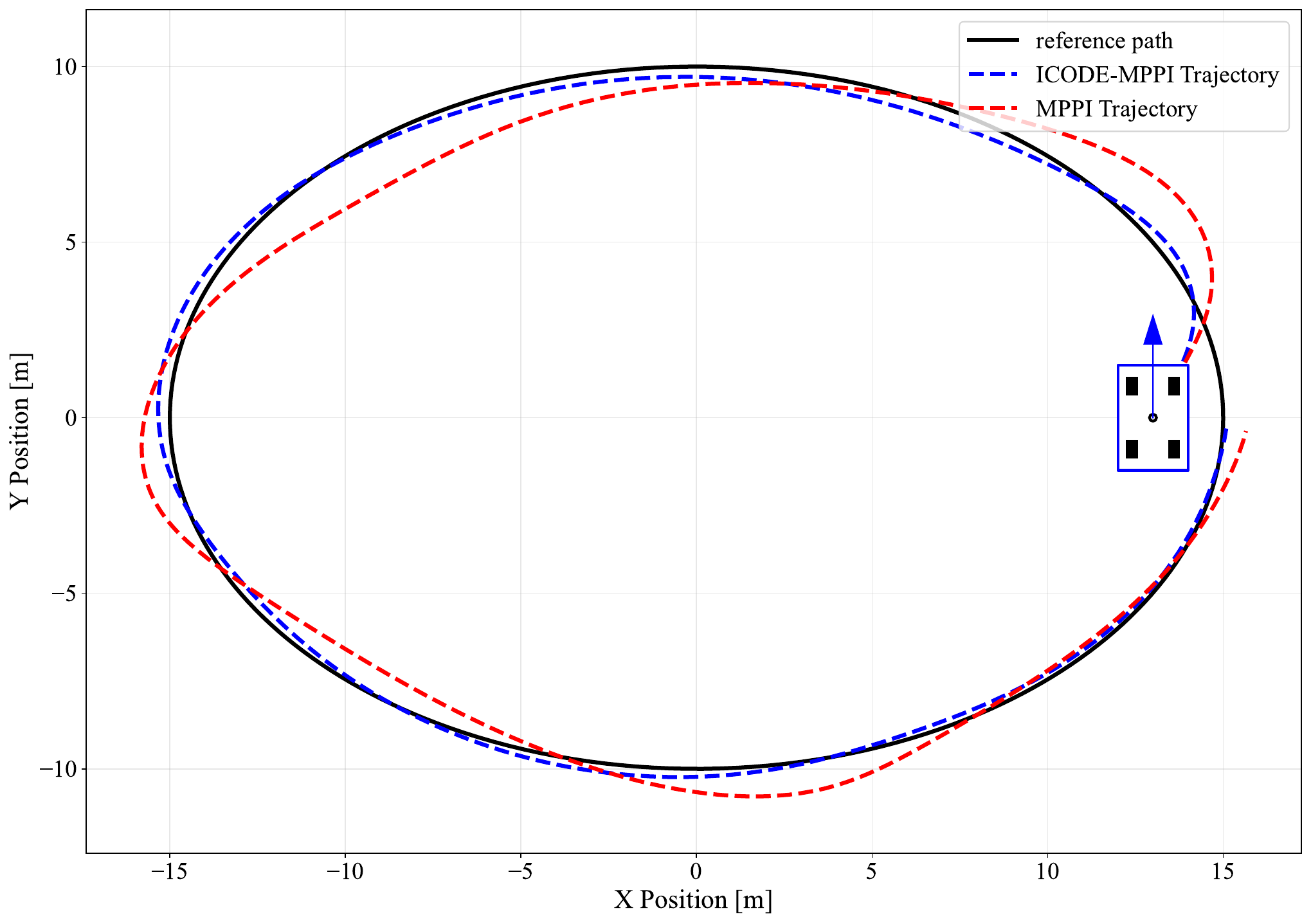}
        \caption{Ellipse Trajectory}
    \end{subfigure}
    \hfill
    \begin{subfigure}[b]{0.32\textwidth}
        \centering
        \includegraphics[width=\textwidth]{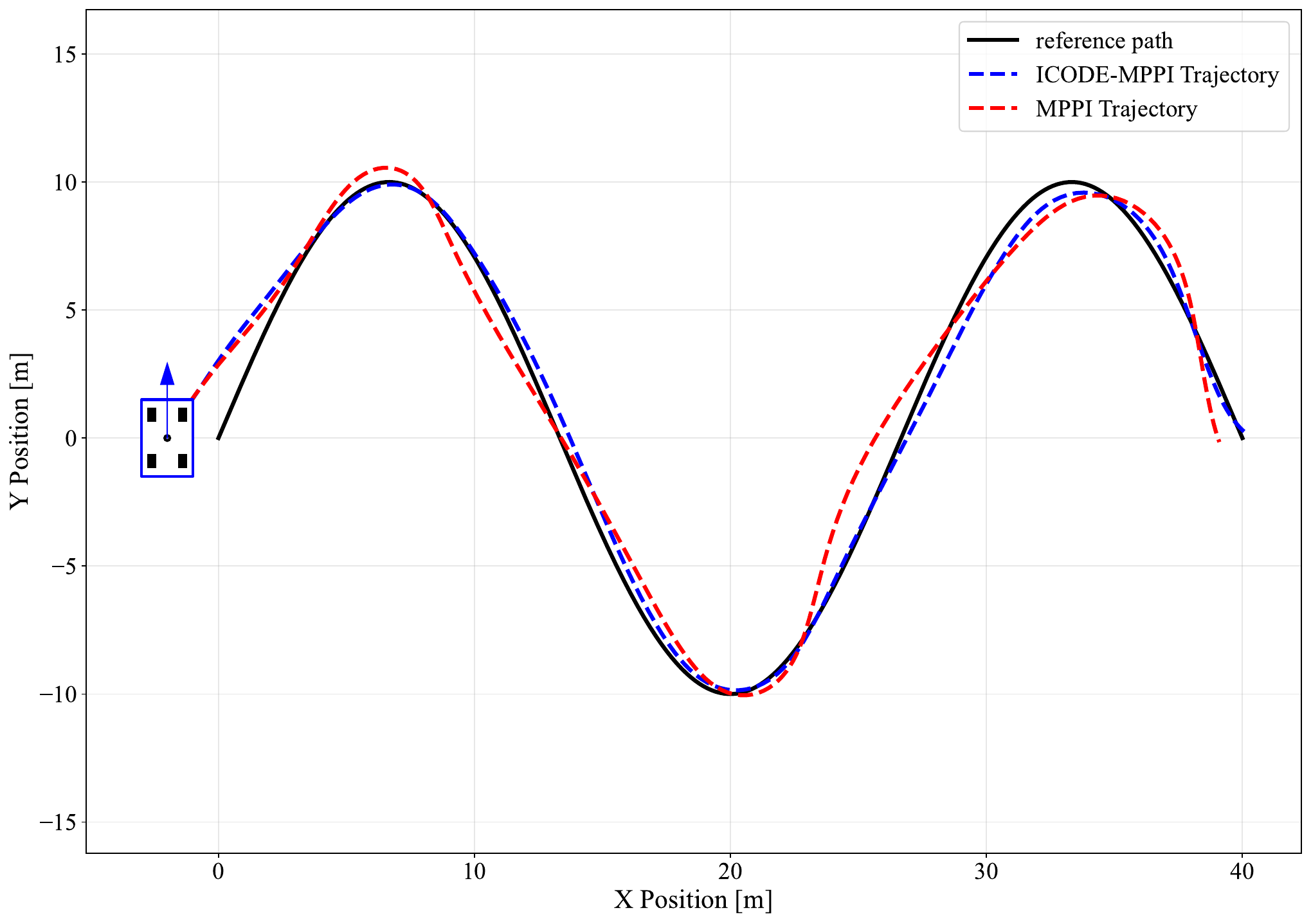}
        \caption{Sine-wave Trajectory}
    \end{subfigure}
    \hfill
    \begin{subfigure}[b]{0.32\textwidth}
        \centering
        \includegraphics[width=\textwidth]{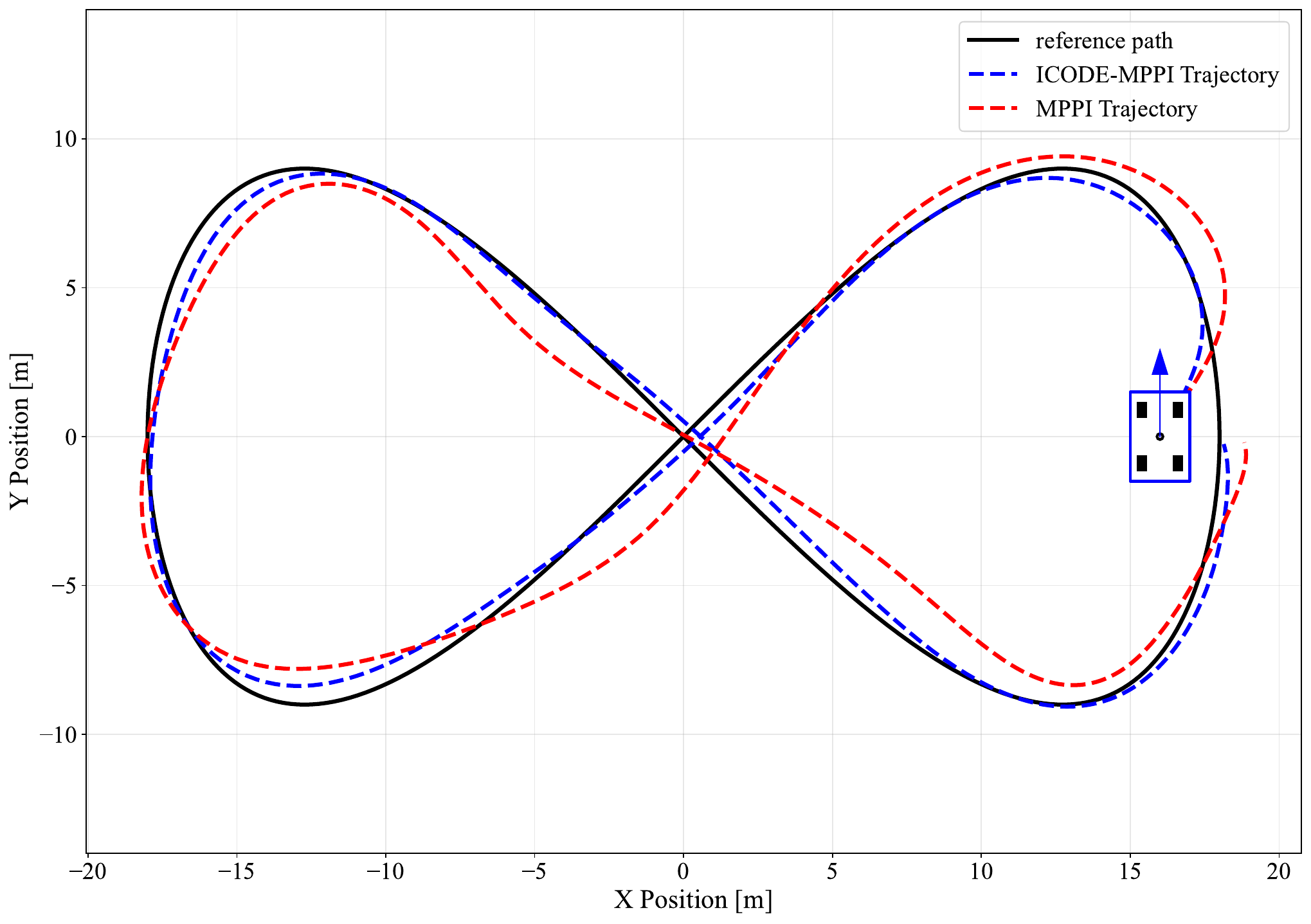}
        \caption{Figure-8 Trajectory}
    \end{subfigure}
    \caption{Trajectory tracking results. The nominal MPPI shows significant steady-state drift under disturbances, particularly at the curvature inflection points, while ICODE-MPPI closely adheres to the reference.}
    \label{fig:trajectories_wide}
\end{figure*}

The spatial tracking performance across the three distinct trajectories is qualitatively evaluated in Fig. \ref{fig:trajectories_wide}. 

As shown in Fig. \ref{fig:trajectories_wide}, the unmodeled time-varying disturbance $\bm{\delta}(t)$ induces a persistent steady-state offset in the nominal MPPI across all paths. However, the severity varies: in the Ellipse, the drift is relatively uniform, pushing the vehicle outward. In contrast, the Figure-8 and Sine-wave paths feature continuous curvature inversions. At these inflection points, the nominal controller struggles to quickly reverse its steering bias, resulting in severe overshooting. 

ICODE-MPPI effectively suppresses this drift. By utilizing the ICODE to continuously approximate the residual dynamics $\bm{\delta}(t)$, the controller achieves a pre-steering effect. It anticipates the lateral push of the disturbance and compensates proactively, maintaining tight adherence to the reference even during sharp, transient maneuvers.

\subsection{Quantitative Analysis and Statistical Robustness}
Table \ref{tab:quantitative} quantifies the tracking precision using Root Mean Square Error (RMSE). 

\begin{table}[htbp]
    \caption{Tracking RMSE Comparison}
    \centering
    \renewcommand{\arraystretch}{1.1}
    \resizebox{\columnwidth}{!}{
        \begin{tabular}{lcccc}
            \toprule
            \textbf{Trajectory} & \textbf{Method} & \textbf{X Error (m)} & \textbf{Y Error (m)} & \textbf{Yaw Error (rad)} \\
            \midrule
            Ellipse & MPPI & 0.589 & 0.557 & 0.231 \\
            & \textbf{ICODE-MPPI} & \textbf{0.448} & \textbf{0.171} & \textbf{0.244} \\
            \hline
            Sine-wave & MPPI & 0.681 & 0.314 & 0.214 \\
            & \textbf{ICODE-MPPI} & \textbf{0.587} & \textbf{0.213} & \textbf{0.266} \\
            \hline
            Figure-8 & MPPI & 0.655 & 0.589 & 0.259 \\
            & \textbf{ICODE-MPPI} & \textbf{0.442} & \textbf{0.216} & \textbf{0.267} \\
            \bottomrule
        \end{tabular}
    }
    \label{tab:quantitative}
\end{table}

The data reveals that ICODE-MPPI achieves a significant reduction in positional errors, with the $Y$-direction error dropping by up to 69\% in the Ellipse case. However, an interesting observation is that the yaw RMSE for ICODE-MPPI is slightly higher than the nominal model. This indicates a control trade-off: to maintain strict positional adherence ($X, Y$) against strong lateral disturbances, the controller performs more aggressive heading adjustments. In path-tracking tasks, prioritizing spatial safety over perfect heading alignment is a deliberate and necessary strategy for robust navigation.

\begin{figure}[!ht]
    \centering
    \begin{subfigure}[b]{1.0\linewidth}
        \centering
        \includegraphics[width=1.0\linewidth]{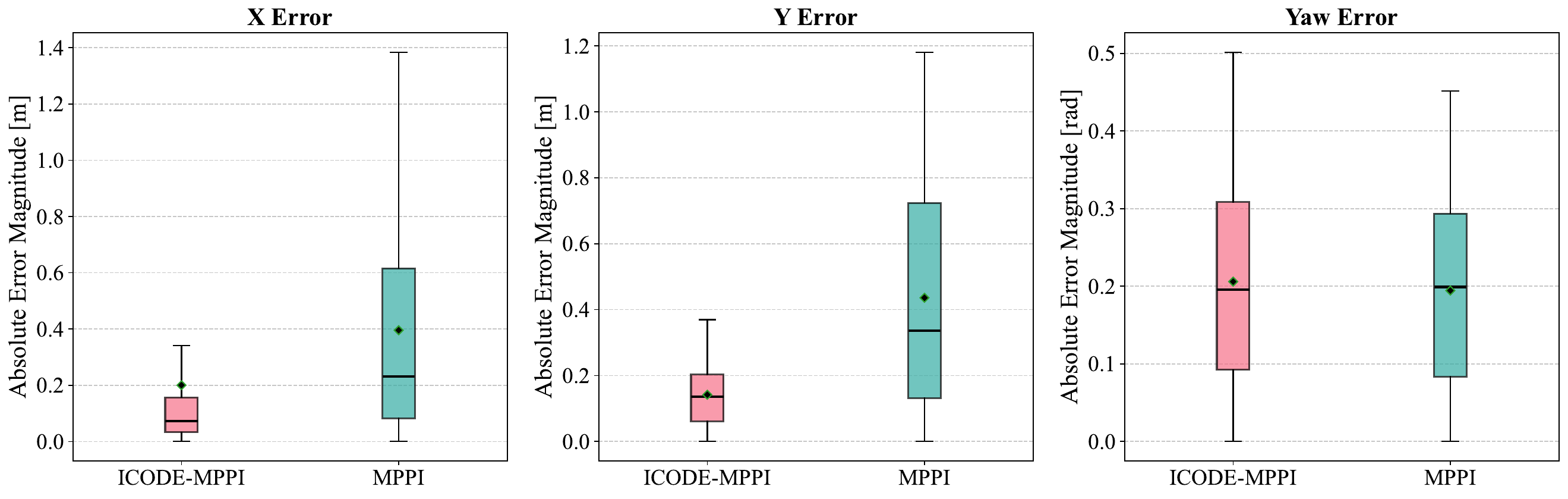}
        \caption{Error Distribution: Ellipse}
    \end{subfigure}
    \vspace{2mm}
    
    \begin{subfigure}[b]{1.0\linewidth}
        \centering
        \includegraphics[width=1.0\linewidth]{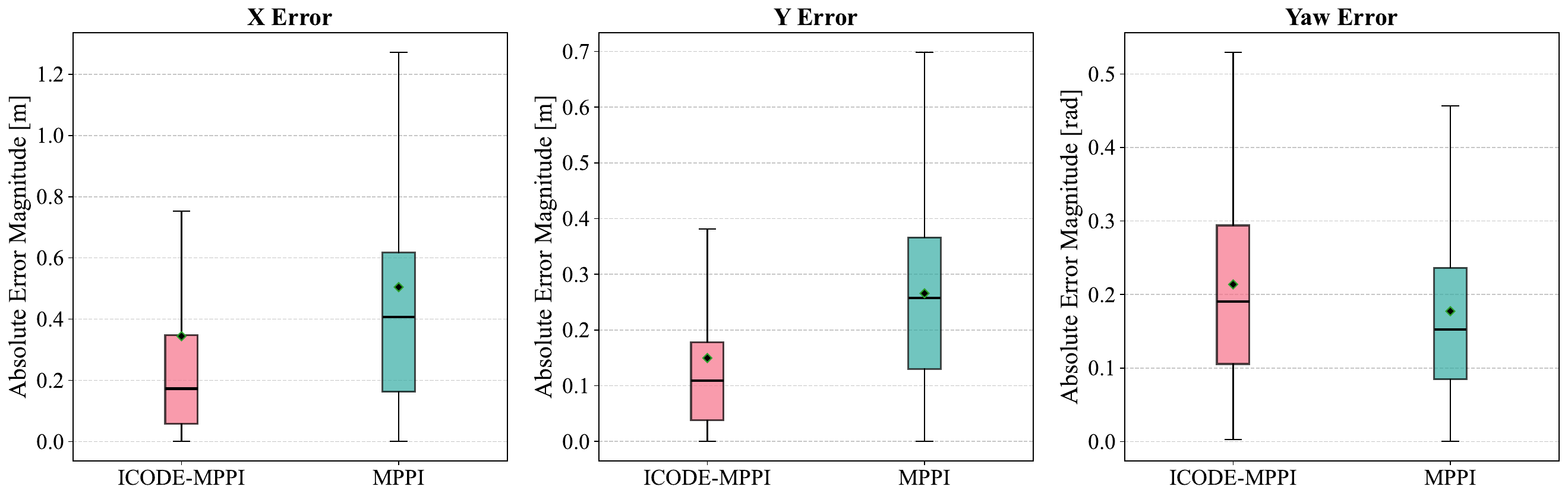}
        \caption{Error Distribution: Sine-wave}
    \end{subfigure}
    \vspace{2mm}
    
    \begin{subfigure}[b]{1.0\linewidth}
        \centering
        \includegraphics[width=1.0\linewidth]{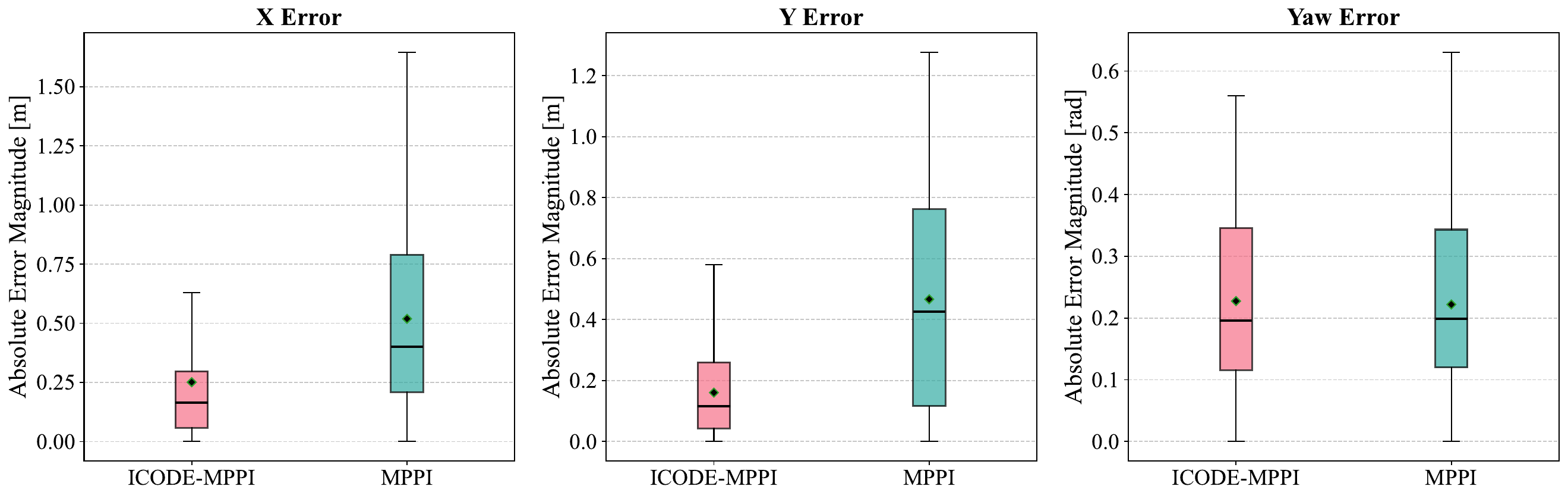}
        \caption{Error Distribution: Figure-8}
    \end{subfigure}
    \caption{Boxplots of state-variable tracking errors ($X, Y, \theta$) across different trajectories.}
    \label{fig:boxplots_single}
\end{figure}

To further analyze the error distribution, Fig. \ref{fig:boxplots_single} presents the absolute error boxplots. In these plots, the \textbf{center line} denotes the median, the \textbf{green point} represents the mean, and the \textbf{whiskers} indicate the range of non-outlier maximum/minimum errors.

The boxplots reveal stark differences in how the errors are distributed. Across all three trajectories, the $X$ and $Y$ positional boxes for ICODE-MPPI are drastically shorter and lower, proving its high resistance to lateral and longitudinal drift. However, the Yaw behavior is more nuanced.  Notably, in the Ellipse and Sine-wave trajectory, the Yaw whiskers for ICODE-MPPI are longer than those of the nominal MPPI. This aligns with our RMSE findings: to maintain strict $X$ and $Y$ coordinates against periodic crosswinds, the ICODE framework occasionally permits marginally larger transient yaw deviations at the inflection points. It sacrifices a small degree of transient heading stability to guarantee absolute positional safety.

\subsection{Control Smoothness and Stability}
\begin{figure}[!ht]
    \centering
    \begin{subfigure}[b]{1.0\linewidth}
        \centering
        \includegraphics[width=1.0\linewidth]{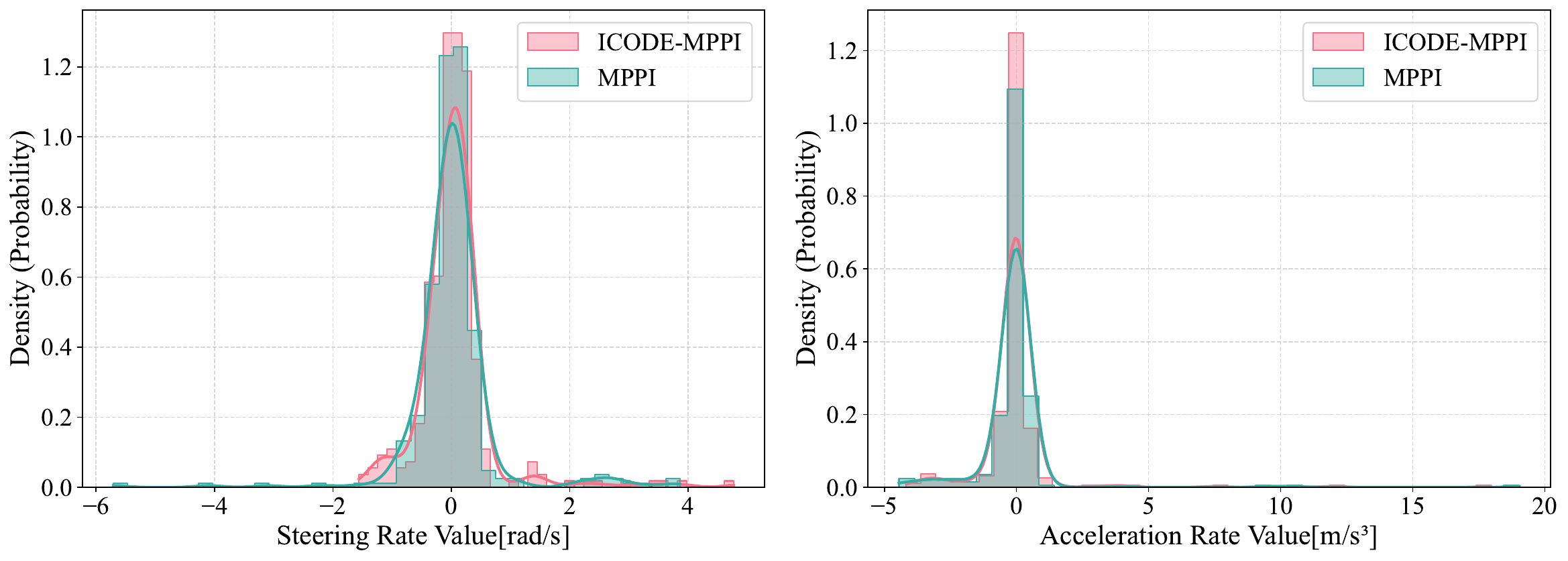}
        \caption{Control Profile: Ellipse}
    \end{subfigure}
    \vspace{2mm}
    
    \begin{subfigure}[b]{1.0\linewidth}
        \centering
        \includegraphics[width=1.0\linewidth]{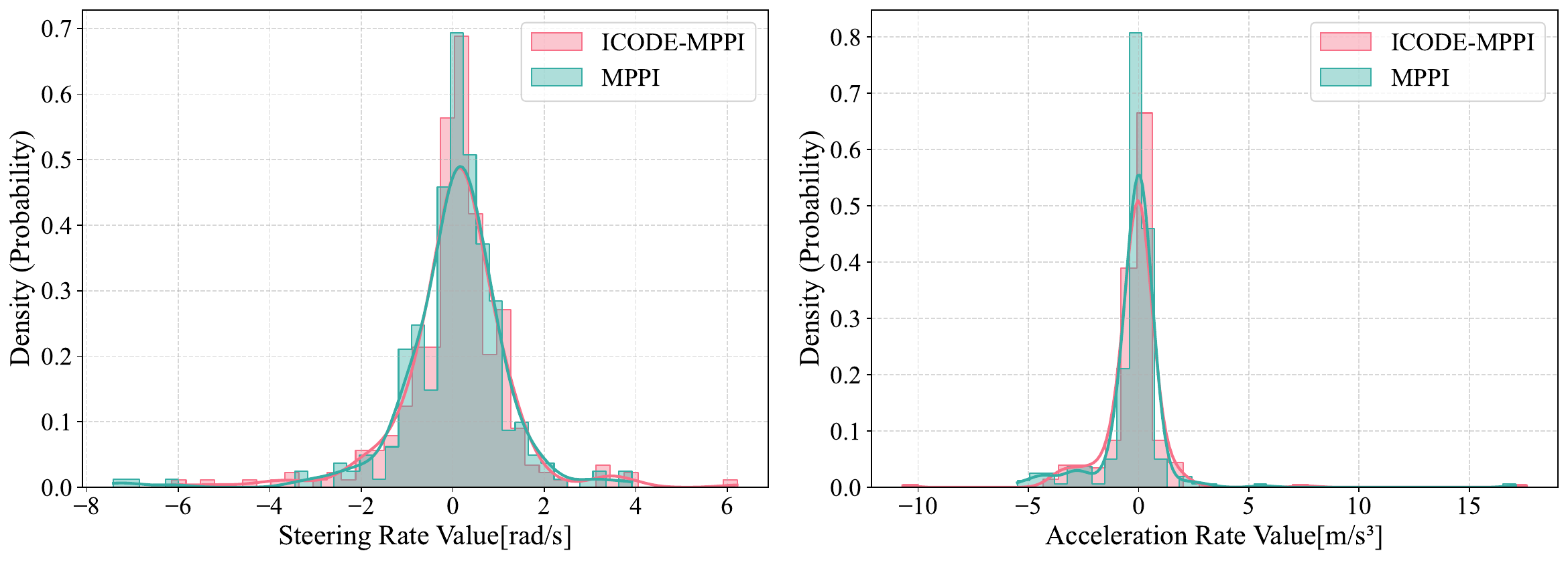}
        \caption{Control Profile: Sine-wave}
    \end{subfigure}
    \vspace{2mm}
    
    \begin{subfigure}[b]{1.0\linewidth}
        \centering
        \includegraphics[width=1.0\linewidth]{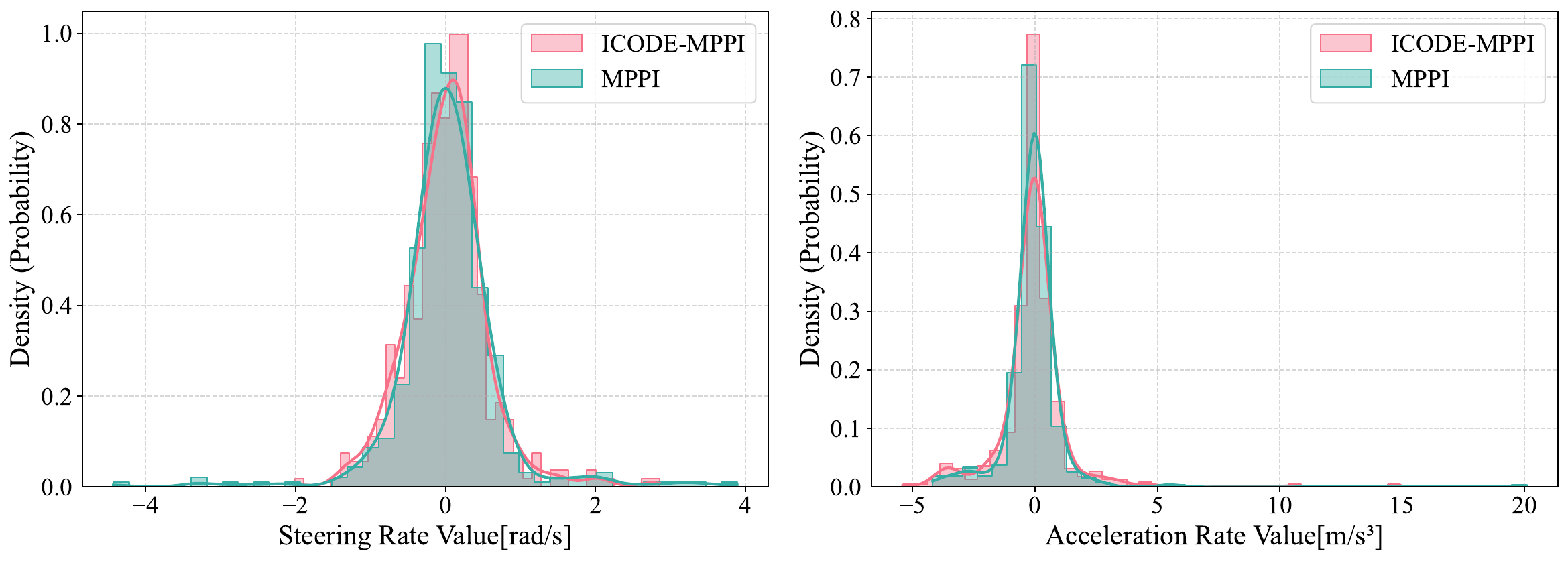}
        \caption{Control Profile: Figure-8}
    \end{subfigure}
    \caption{Probability density of control input rates. ICODE-MPPI drastically smooths the steering commands, with acceptable trade-offs in longitudinal acceleration distribution.}
    \label{fig:smoothness_single}
\end{figure}

For real-world deployment, the control effort must be feasible and smooth. Fig. \ref{fig:smoothness_single} illustrates the probability density of the control jerk (rate of change for acceleration $a$ and steering $\delta$). The key metrics are the peak value at zero (representing steady-state periods) and the distribution width (representing chattering intensity).Higher peaks indicate enhanced stability, while narrower distributions signify reduced chattering.

In all scenarios, the nominal MPPI exhibits a flat, wide distribution for steering rate, indicating severe control chattering.It constantly reacts to the disturbance with sudden, large-magnitude steering inputs.The higher peak at zero and the narrower distribution of ICODE-MPPI effectively demonstrate its superior stability and reduced chattering compared to the nominal MPPI.

However, a careful examination reveals an anomaly in certain trajectories (e.g., Sine-wave): while the steering rate is vastly improved, ICODE-MPPI's peak is lower, and the distribution is slightly wider than the nominal case in the acceleration rate plots. This is physically justifiable due to lateral-longitudinal coupling. By maintaining aggressive and precise steering corrections to perfectly track the complex curves under heavy disturbances, the vehicle inevitably experiences higher lateral tire forces, causing slight losses in forward momentum. Consequently, the longitudinal acceleration controller must actuate more frequently to maintain the reference speed profile. Trading a slight increase in longitudinal throttle adjustments for a massive gain in lateral safety and steering smoothness is highly desirable in autonomous driving applications.

\section{Conclusion}
This paper presented ICODE-MPPI, a robust path-tracking framework that addresses the model mismatch challenge by embedding Input Concomitant Neural ODEs as residual learners. By explicitly modeling the residual terms of the system's state-space equations, our method achieves superior accuracy in tracking positional states ($X, Y$) under high-disturbance conditions. The simulation results across multiple complex trajectories confirm that ICODE-MPPI significantly reduces state-variable drift, while maintaining a strategic trade-off with transient heading alignment. Moreover, the analysis of control effort highlights that while steering chattering is drastically suppressed, the longitudinal-lateral coupling requires slightly more active acceleration adjustments to maintain reference velocities. 

Future work will focus on extending the framework to incorporate dynamic obstacle avoidance capabilities and validating the system's real-time performance through deployment on physical scale-model vehicles.

\bibliographystyle{IEEEtran}
\bibliography{refs}

\end{document}